\pdfoutput=1 

\documentclass[11pt]{article}

\usepackage[final]{EMNLP2022}

\usepackage{times}
\usepackage{latexsym}
\usepackage{subcaption}
\usepackage{amsmath,amsfonts,amssymb}
\usepackage{breqn}
\usepackage{tabularx}
\usepackage{multirow}
\usepackage{graphicx}
\usepackage{color}
\usepackage{tcolorbox}
\usepackage{CJK}
\usepackage{adjustbox}
\usepackage{xcolor}
\usepackage{colortbl}
\usepackage{multicol}
\usepackage{vwcol} 
\newsavebox{\algleft}
\newsavebox{\algright}
\usepackage{bm}
\usepackage{booktabs}
\usepackage{ctable} 
\usepackage{amsmath,stackengine}
\usepackage[linesnumbered,ruled,vlined]{algorithm2e}
\usepackage{times}
\usepackage{latexsym}
\usepackage{bbm}
\usepackage[utf8]{inputenc} 
\usepackage[T1]{fontenc}    
\usepackage{hyperref}       
\usepackage{url}            
\usepackage{booktabs}       
\usepackage{amsfonts}       
\usepackage{nicefrac}       
\usepackage{microtype}      
\usepackage{xcolor}         
\usepackage{bm}
\usepackage{microtype}
\newcommand{\zixuan}[1]{{\color{blue}{\small\bf\sf [zixuan: #1]}}}

\usepackage[T1]{fontenc}

\usepackage[utf8]{inputenc}

\usepackage{microtype}

\usepackage{inconsolata}

\title{Adapting a Language Model While Preserving its General Knowledge}

\author{
Zixuan Ke$^{1}$, Yijia Shao$^{2}$, Haowei Lin$^{2}$, Hu Xu$^{3}$, Lei Shu$^{1}$\thanks{~~Now at Google Research \url{leishu@google.com}}~~~and Bing Liu$^{1}$\\ 
$^1$Department of Computer Science, University of Illinois at Chicago\\
$^2$Wangxuan Institute of Computer Technology, Peking University\\
$^3$Meta AI\\
$^1$\texttt{\{zke4,liub\}@uic.edu}\\  $^2$\texttt{\{shaoyj, linhaowei\}@pku.edu.cn} \\ $^3$\texttt{huxu@fb.com} \\
}

\begin{document}

\maketitle

\begin{abstract}
Domain-adaptive pre-training (or \textit{DA-training} for short), also known as \textit{post-training}, aims to train a pre-trained general-purpose language model (LM) using an unlabeled corpus of a particular domain to adapt the LM so that end-tasks in the domain can give improved performances. However, existing DA-training methods are in some sense \textit{blind} as they do not explicitly identify what knowledge in the LM should be preserved and what should be changed by the domain corpus. 
This paper shows that the existing methods are sub-optimal and proposes a novel method to perform \textit{a more informed adaptation} of the knowledge in the LM by (1) soft-masking the attention heads based on their importance to best preserve the general knowledge in the LM and (2) contrasting the representations of the general and the full (both general and domain knowledge) to learn an integrated representation with both general and domain-specific knowledge. Experimental results will demonstrate the effectiveness of the proposed approach.\footnote{{\color{red}\url{https://github.com/UIC-Liu-Lab/DGA}}}

\end{abstract}

\section{Introduction}
\label{sec.intro}

Pre-trained general-purpose language models (LMs) like BERT~\cite{DBLP:conf/naacl/DevlinCLT19}, RoBERTa~\cite{DBLP:journals/corr/abs-1907-11692}, and GPT-3~\cite{brown2020language} have become a standard component in almost all NLP applications. 
Researchers have also found that \textit{domain-adaptive pre-training} (or \textit{DA-training} for short) using an \textit{unlabeled} corpus in a specific domain to adapt an LM can further improve the end-task performance in the domain
\cite{DBLP:conf/acl/GururanganMSLBD20,DBLP:conf/naacl/XuLSY19,xu2019review,sun2019fine,alsentzer2019publicly}.~Note that \textit{domain-adaptive pre-training} is also called \textit{post-training}~\cite{DBLP:conf/naacl/XuLSY19}.  

Existing DA-training methods 
simply apply the \textit{same} pre-training objective, i.e., the \textit{mask language model} (MLM) loss, to further train an LM using a domain corpus. These methods are sub-optimal because they do not explicitly identify what should be preserved and what should be updated in the LM by the domain corpus. 

This paper argues that a good DA-training method has two needs. On the one hand, the general language knowledge learned in the LM should be preserved as much as possible because the target domain data is typically not large enough to be sufficient to learn the general knowledge well. For example, some words and their contexts may appear infrequently in a particular domain. The knowledge about them cannot be learned accurately based on the domain data alone. When these words and contexts appear in an end-task, the system will have difficulties. Thus, we need to rely on the knowledge about them in the LM. Since existing DA-training updates the LM with little guidance, such useful general knowledge may be corrupted. 
On the other hand, due to polysemy (same word with different meanings in different domains) and the fact that different domains also have their special word usages and contexts, the LM should be specialized or adapted to the target domain. A good DA-training should balance these two needs to adapt the LM to the target domain with minimal corruption 
to the good general knowledge in the LM. 

This paper proposes a novel technique to enable a more \textit{informed adaptation} to 
(1) preserve the \textit{general} knowledge in the LM as much as possible, and (2) update the LM to incorporate the \textit{domain-specific} knowledge of the target domain as needed. The focus of the existing DA-training research has been on (2). As we argued above, (1) is also important as focusing only on (2) may destroy some useful general knowledge and produce sub-optimal results for end-tasks.
To achieve (1), the system should constrain the gradient update of each attention head\footnote{We will see in Sec.~\ref{sec:experiments} that constraining the neurons in other layers is unnecessary.} based on its importance to the \textit{general knowledge} so that the {general knowledge} in LM can be preserved as much as possible. With (1), (2) will be able to 
change the part of the general knowledge that needs to be updated to adapt the LM to suit the target domain.\footnote{This is very different from \textit{continual learning} (CL)~\cite{chen2018lifelong} as CL needs to preserve the past knowledge to deal with \textit{catastrophic forgetting}~\cite{mccloskey1989catastrophic}. DA-training can and should change/adapt the general knowledge in the original LM to suit the target domain. }

In this paper, we propose a novel model called DGA (\textit{\textbf{D}A-training - \textit{\textbf{G}}eneral knowledge preservation and LM \textbf{\textit{A}}daptation}) for the purpose. {\color{black}The key idea of the proposed method is to preserve the general language knowledge in the LM while adapting the LM to a specific domain. However, it is not obvious how this can be done, i.e., how to find those parameters that are important for the general knowledge and how to protect them. This paper proposes a novel proxy-based method to achieve the objectives. It works as follows. DGA first estimates the importance of each attention head in the LM via the newly proposed \textit{proxy KL-divergence loss} 
(Sec.~\ref{sec: prevent_forgetting}). This importance score reflects how important each attention head is to the general knowledge.}
Based on the importance scores, it performs two key functions: The first function 
uses the scores to \textit{soft-mask} (rather than binary-mask or completely block) the gradient update to prevent important general knowledge in LM from being unnecessarily corrupted. This is related to \textit{pruning of unimportant attention heads}~\cite{michel2019sixteen}. However, pruning is not directly applicable to DA-training as we will show in Sec.~\ref{Sectionrelated.work}. The proposed soft-masking constrains only the backward gradient flow in training. It is not necessary to soft-mask the forward pass in either training or inference. This is important because using the knowledge in the full network encourages maximal integration of pre-trained general knowledge and the target domain-specific knowledge. 
{\color{black}The second function contrasts the representation for the general knowledge in the LM and the full (including both the general and the domain-specific) knowledge to learn an integrated representation (Sec.~\ref{sec.contrast}).\footnote{{\color{black}Contrasting the general and only the domain-specific knowledge gives poorer results (see Sec.~\ref{sec:results}) as it causes the two types of knowledge to split rather than to integrate.} }
}  

In summary, this paper makes two key contributions. 

\textbf{(1).} 
    It proposes the idea of \textit{informed adaptation} to integrate the specialized knowledge in the target domain into the LM with minimal corruption to the useful general knowledge in the original LM. 
    
    \textbf{(2).} It proposes a new model DGA with two novel functions to enable better DA-training. DGA estimates the attention head importance to protect the important general knowledge in the LM and integrates the specialized knowledge in the target domain into the LM through contrasting the general and the full knowledge. 
    
    To the best of our knowledge, none of these has been reported in the literature before. 
    
    Extensive experiments have been conducted in 6 different domains and on 10 baselines to demonstrate the effectiveness of the proposed DGA. 

\section{Related Work}
\label{Sectionrelated.work}

\vspace{+1mm}
\noindent
\textbf{Domain-adaptive pre-training (DA-training).}
Researchers have applied DA-training to many domains, e.g., reviews \cite{DBLP:conf/naacl/XuLSY19,xu2019review}, biomedical text \cite{lee2020biobert}, 
news and papers \cite{DBLP:conf/acl/GururanganMSLBD20}, and social media \cite{chakrabarty2019imho}. 
However, they all use the same mask language model (MLM) loss. {\color{black}We argue that it is sub-optimal and it is also important to preserve the general knowledge in the LM as much as possible and integrate it with the target domain knowledge.} 

\vspace{+1mm}
\noindent
\textbf{Network pruning as importance computation.}
{\color{black}It is known that many parameters in a neural network are redundant and can be pruned~\cite{li2021differentiable,lai2021parp}.
This has also been shown for pre-trained Transformer~\cite{chen2020lottery,lin2020pruning,gao2021adapting,michel2019sixteen,voita2019analyzing}. 
A popular pruning method is to discard the parameters with small absolute values~\cite{han2015learning,guo2016dynamic}. 
Other methods prune the network at a higher level. In a Transformer-based model, these include pruning the attention head \cite{michel2019sixteen,voita2019analyzing,mccarley2019structured} and pruning sub-layers in a standard Transformer layer \cite{Fan2020Reducing,sajjad2020effect}. 
However, the above methods are not directly applicable to us as we need to compute the head importance for the {LM} using unlabeled domain data, while the above approaches are all for \textit{supervised end-tasks}. We propose to use a \textit{proxy KL-divergence} loss for our purpose. {\color{black} Note that it is possible to prune other sub-layers in the Transformer. However, as shown in Sec. \ref{sec:results}, estimating the importance for other layers does not improve the performance.}
}

\vspace{+1mm}
\noindent
\textbf{Contrastive learning.}
Contrastive learning~\cite{chen2020simple,he2020momentum} can learn good 
representations by maximizing the similarity of positive pairs and minimizes that of negative pairs:
\begin{equation}
\label{eq.relate_contrast}
    \mathcal{L}_{\text{contrast}} = -\log\frac{e^{(\text{sim}(q_i, q^+_i)/\tau)}}{\sum_{j=1}^Ne^{(\text{sim}(q_i,q^+_j)/\tau)}},
\end{equation}
{\color{black}where $N$ is the batch size, $\tau$ is a temperature parameter, $\text{sim}(\cdot)$ is a similarity metric, and $q_i$ and $q^+_i$ are representations for positive pairs $x_i$ and $x_i^+$ (typically, $x_i^+$ is an augmented sample of $x_i$, {\color{black}e.g., generated via cropping, deletion or synonym replacement \cite{gao2021simcse}}). In the \textit{unsupervised contrastive loss}, the negative samples are the other samples in the batch,  
indicated in the denominator.}

We mainly use contrasive loss to contrast the representations of the important general knowledge in the original LM and the full knowledge (both the general and domain-specific knowledge) to achieve a good integration of the general knowledge and the domain specific knowledge. 

\section{Proposed DGA System}
\label{Sec: preliminary}



\begin{figure}[t!]
\centering
\includegraphics[width=\columnwidth]{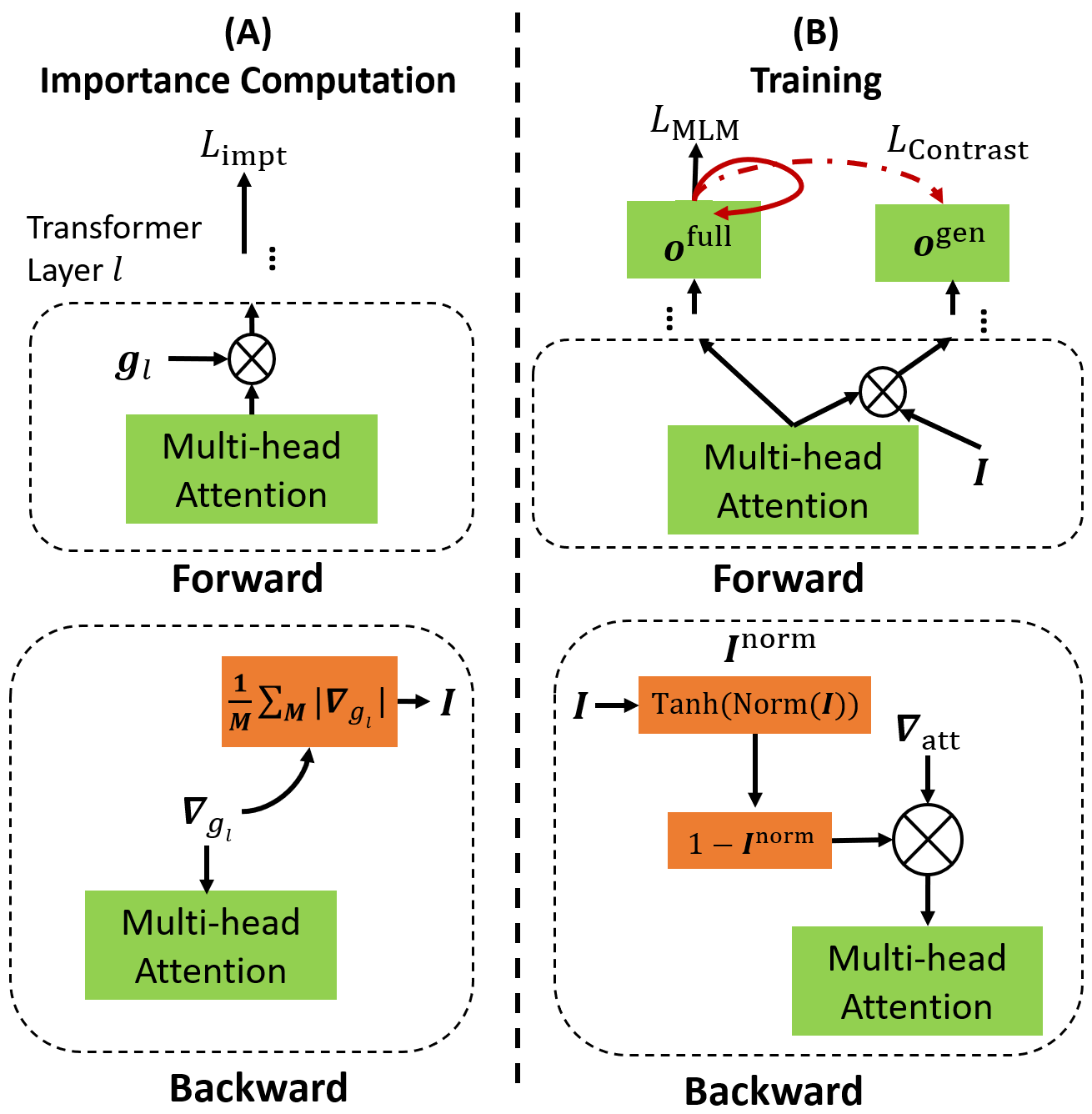}
\vspace{-6mm}
\caption{
Illustration of DGA. {\color{black}\textbf{(A)} shows the importance computation. This is done by adding a gate vector $\bm{g}_l$ multiplying with the multi-head attention (Eq.~\ref{eq:gmhatt}) and averaging its training gradients (Eq.~\ref{eq:importance}). 
\textbf{(B)} shows DGA training. 
In backward pass, attention heads are soft-masked based on their importance $\bm{I}$ (Eqs. \ref{eq:mask} and \ref{eq:norm}) to try to preserve the general knowledge in the LM as much as possible. 
In forward pass, the added gate vector is removed except for feature learning in the contrastive loss. 
The contrastive loss is computed by contrasting the general knowledge with importance ($\bm{o}^{\text{gen}}$ in Eq.~\ref{eq:final_general_knowledge}) applied and the full knowledge without {\color{black}applying the importance ($\bm{o}^{\text{full}}$ in Eq.~\ref{eq:final_specialized_knowledge})}.} The final objective of DGA consists of MLM loss and contrastive loss. Note that we omit the details of other parts of Transformer and only focus on the multi-head attention mechanism.
}
\label{fig:overview}
\vspace{-3mm}
\end{figure}

As discussed earlier, DGA goes beyond the MLM loss to perform two more functions: (1) preserving the important general knowledge in the LM by soft-masking the attention heads based on their importance. This helps avoid potential corruption of the general knowledge in the LM in DA-training (Sec.~\ref{sec: prevent_forgetting}). {\color{black}However, the challenge is how to identify the general knowledge in the LM and how to protect it. We will propose a method to do that.} (2) encouraging the model to learn integrated representations of the target domain and the general knowledge in the LM 
(Sec.~\ref{sec.contrast}). {\color{black}It is also not obvious how this can be done. We propose a contrastive learning based method to do it.}  Figure~\ref{fig:overview} gives an overview of DGA. 


\subsection{Preserving General Knowledge by Soft-Masking Attention Heads}
\label{sec: prevent_forgetting}

\textbf{Multi-head attention.} Multi-head attention is arguably the most important component in the Transformer model \cite{vaswani2017attention}. We omit details of other parts and refer the reader to the original paper. Formally, let $\bm{x}=x^{(1)},...,x^{(T)}$ be a sequence of $T$ real vectors where $x^{(t)}\in\mathbb{R}^d$ and let $q\in\mathbb{R}^d$ be a query vector.
The \textit{attention mechanism} is defined as 
\begin{equation}
\text{att}(\bm{x},q) = W_o\sum_{t=1}^T\alpha^{(t)}(q)W_vx^{(t)},
\end{equation}
where
\begin{equation}
\alpha^{(t)}(q)=\text{softmax}(\frac{q^TW_q^TW_kx^{(t)}}{\sqrt{d}}).
\end{equation}
The projection matrices $W_o, W_v, W_q, W_k\in\mathbb{R}^{d\times d}$ are learnable parameters. The query vector is from the same sequence as $\bm{x}$ in self-attention. A Transformer contains $L$ identical layers. For layer $l$, $H_l$ different attention heads are applied in parallel to enable the Transformer to be trained on more data. Simply put, \textit{multi-head attention} (mhatt) is the simultaneous application of multiple attention heads in a single Transformer architecture. They are then applied in parallel to obtain multi-head attention.\footnote{We follow the notation in \cite{michel2019sixteen}, where the notation in Eq.~\ref{eq:mhatt} is equivalent to the ``concatenation'' formulation in \cite{vaswani2017attention}.}
\begin{equation}
\label{eq:mhatt}
\text{mhatt}_l(\bm{x},q)=\sum_{h=1}^{H_l}\text{att}_{lh}(\bm{x},q),
\end{equation}
where $h$ indicates the $h^{th}$ attention head. Note that the input $\bm{x}$ is different in each layer since the input of a given layer is the output of last layer. To ease the notation, we use the input $\bm{x}$ for all layers.

\textbf{Head importance.} Researchers have found that not all attention heads are important \cite{michel2019sixteen}. We introduce a \textit{gate vector}, $\bm{g}_{l}$, where each cell is a \textit{gate variable}, $g_{lh}$, to the attention head summation for detecting the importance of attention heads. The resulting importance scores are used to soft-mask the  heads in DA-training.
\begin{equation}
\label{eq:gmhatt}
\text{gmhatt}_l(\bm{x},q)=\sum_{h=1}^{H_l}g_{lh}, \otimes\text{att}_{lh}(\bm{x},q)
\end{equation}
where $\otimes$ is the element-wise multiplication. A gradient-based head importance detection method is proposed in \cite{michel2019sixteen}. Given a dataset $D=\{(\bm{y}_m,\bm{x}_m)\}_{m=1}^M$ of $M$ samples  ($\bm{y}_m$ is the label of $\bm{x}_m$ as \citet{michel2019sixteen} worked on supervised learning), the importance of a head is estimated with a gradient-based proxy score
\begin{equation}
\label{eq:importance}
I_{lh} = \frac{1}{M}\sum_{m=1}^M|\nabla_{g_{lh}}|,
\end{equation}
where $\nabla_{g_{lh}}$ is the gradient of gate variable $g_{lh}$,
\begin{equation}
\label{eq:gradient}
\nabla_{g_{lh}} = \frac{\partial\mathcal{L}_{\text{impt}}(\bm{y}_m,\bm{x}_m)}{\partial g_{lh}},
\end{equation}
where $\mathcal{L}_{\text{impt}}$ is a task-specific/domain-specific loss function. The gradient can be used as the importance score because changing ${g_{lh}}$ is liable to have a large effect on the model if $I_{lh}$ has a high value. 

Although Eq.~\ref{eq:importance} offers a way to compute the importance of attention heads \textit{w.r.t. a given loss $\mathcal{L}_{\text{impt}}$}, we are unable to directly apply it: If we use the domain data at hand and the MLM loss as $\mathcal{L}_{\text{impt}}$, $\nabla_{g_{lh}}$ only indicates the importance score for \textit{domain-specific} knowledge. However, our goal is to estimate the attention heads importance for the \textit{general knowledge} in LM which requires the data used in training the LM to compute the $\mathcal{L}_{\text{impt}}$. In practice, such data is not accessible to users of the LM. Further, label is needed in Eq.~\ref{eq:importance} but our domain corpus is unlabeled in DA-training. 
To address these issues, we propose to compute a \textit{proxy KL-divergence loss} for $\mathcal{L}_{\text{impt}}$.

{\color{black}\textbf{Proxy KL-divergence loss.}} 
We need a proxy for $\mathcal{L}_{\text{impt}}$ such that its gradient ($\nabla_{g_{lh}}$)
can be used to compute head importance without using the LM's original pre-training data. We propose to use model \textit{robustness} as the proxy, i.e., we try to detect heads that are important for LM's \textit{robustness}. Its gradient, $\nabla_{g_{lh}}$, then indicates the robustness and thus the importance to the LM model. Our rationale is as follows: If an $I_{lh}$ (the average of $|\nabla_{g_{lh}}|$, see Eq.~\ref{eq:importance}) has a high value, it indicates that it is important to the LM's robustness because its change can cause the LM to change a great deal. It is thus an important head to the LM. In contrast, if $I_{lh}$ has a small value, it is a less or not important head to the LM.

To compute the \textit{robustness} of the LM, we take a subset (a hyper-parameter) of the target domain data $\{\bm{x}^{\text{sub}}_m\}$ (no label in DA-training) and input $\bm{x}^{\text{sub}}_m$ twice to the LM and compute the KL-divergence of the two resulting representations,
\begin{equation}
\label{eq:proxy}
\mathcal{L}_{\text{impt}} = \text{KL}(f_1(\bm{x}^{\text{sub}}_m),f_2(\bm{x}^{\text{sub}}_m)),
\end{equation}
{\color{black}where $f_1$ and $f_2$ are the LM with different dropout masks.} Note that we don't need to add any additional dropouts to implement $f$ because independently sampled \textit{dropout masks} are used as input in the Transformer. In training a Transformer, there are dropout masks placed on fully-connected layers and attention probabilities. Thus, simply feeding the same input to the Transformer twice will get two representations with different dropout masks. Since dropout is similar to adding noise, the difference between the two representations can be regarded as the \textit{robustness} of the Transformer model. {\color{black}Figure \ref{fig:overview} (A) shows how we compute the importance of each attention head using the gradient of the gate vector $\bm{g}_{l}$.}

\textbf{Soft-masking attention heads in DA-training.} Recall we want to preserve the general knowledge in the LM during DA-training using head importance $I_{lh}$. Given the attention head $\text{att}(\bm{x},q)$ and DA-training loss $\mathcal{L}_{\text{DA-train}}$ (typically the MLM loss; we also propose an additional loss in Sec.~\ref{sec.contrast}), we can ``soft mask'' its corresponding gradient ($\nabla_{\text{att}_{lh}}$~\footnote{$\nabla_{\text{att}_{lh}}$ indicates the gradient for attention head $\text{att}_{lh}(\bm{x},q)$, distinguished from $\nabla_{g_{lh}}$ in Eq.~\ref{eq:importance} which is the gradient for the gate variable $g_{lh}$}) using the head importance value $I_{lh}$,
\begin{equation}
\label{eq:mask}
\nabla'_{\text{att}_{lh}} = (1-I^{\text{norm}}_{lh}) \otimes \nabla_{\text{att}_{lh}}, 
\end{equation}
where $I^{\text{norm}}_{lh}$ is from $I_{lh}$ via normalization
\begin{equation}
I^{\text{norm}}_{lh} = |\text{Tanh}(\text{Normalize}(I_{lh}))|.
\label{eq:norm}
\end{equation}
{\color{black} $\texttt{Normalize}$ makes the $I_{lh}$ have a mean of 0 and standard deviation of 1. Absolute value of $\texttt{Tanh}$ ensures that $I_{lh}$ takes values in the interval $[0,1]$. } Eq.~\ref{eq:mask} means to constrain the gradient of the corresponding head $\text{att}_{lh}(\bm{x},q)$ by element-wise multiplying one minus the head importance to the gradient. It is ``soft-masking'' because $I_{lh}$ is a real number in $[0,1]$ (instead of binary \{0, 1\}), which gives the model the flexibility to adjust the attention head. This is useful because although some heads are important to the LM, they may conflict with the knowledge in the target domain and thus need adjusting. {\color{black}Also note that the soft masks here affect only the backward pass and are not used in forward pass (so that forward pass can use the full network and encourage maximal integration of pre-trained general and domain-specific knowledge) except for feature learning using contrastive learning (see below).} 
{\color{black}Figure~\ref{fig:overview} (B) shows that attention heads are soft-masked during training.} 

\subsection{Contrasting General and Full Knowledge}
\label{sec.contrast}

We now present how to integrate the general knowledge in the LM and the domain-specific knowledge in the target domain by contrasting the general knowledge and the full knowledge (both general and domain-specific). We first introduce how we obtain such knowledge 
from the LM for the input $\bm{x}$, and then discuss how we contrast them. 

\textbf{Obtaining the general knowledge for the input sequence $\bm{x}$ from the LM} is by extracting the representation of combining the attention heads and their importance scores ($I^{\text{norm}}_{lh}$ in Eq.~\ref{eq:norm}) in the forward pass. The intuition is that since the importance scores show how important each attention head is to the general knowledge, the resulting representation reflects the main general knowledge used by $\bm{x}$. Formally, we plug $I^{\text{norm}}_{lh}$ (soft-masks) as the gate variable $g_{lh}$ in Eq.~\ref{eq:gmhatt}, 
\begin{equation}
\label{eq:general_knowledge}
\text{gmhatt}^{\text{gen}}_l(\bm{x},q)=\sum_{h=1}^{H_l}I^{\text{norm}}_{lh} \otimes\text{att}_{lh}(\bm{x},q).
\end{equation}
Given the attention heads for general knowledge, we can plug it into the whole Transformer to obtain the final general knowledge (taking the average of each token's output in the input sequence).
\begin{equation}
\label{eq:final_general_knowledge}
\bm{o}^{\text{gen}} = \text{Transformer}(\text{gmhatt}^{\text{gen}}(\bm{x},q)).
\end{equation}
{\color{black}(See $\bm{o}^{\text{gen}}$ also in Figure~\ref{fig:overview} (B)).} 

{\color{black}\textbf{Obtaining the full (both general and domain-specific) knowledge in $\bm{x}$} is similar. The only difference is that we extract the representation of $\bm{x}$ without applying the importance (soft-masks) on attention heads in the forward pass, 
\begin{equation}
\label{eq:specialized_knowledge}
\text{gmhatt}^{\text{full}}_l(\bm{x},q)=\sum_{h=1}^{H_l}\text{att}_{lh}(\bm{x},q).
\end{equation}
Similarly, we can plug it into the Transformer,
\begin{equation}
\label{eq:final_specialized_knowledge}
\bm{o}^{\text{full}} = \text{Transformer}(\text{gmhatt}^{\text{full}}(\bm{x},q)).
\end{equation}
(See $\bm{o}^{\text{full}}$ also in Figure~\ref{fig:overview} (B)). {\color{black}Note that it is possible to use $(1-I_{lh}^{\text{norm}})$ as the importance of domain-specific knowledge and contrast it with the general knowledge. However, this produces poorer results (see Table~\ref{tab:ablation}) as explained in footnote 4. }


\textbf{Contrasting general and full knowledge.} 
It is known that contrastive learning helps learn a good isotropic representation that is good for down-stream tasks, with the help of positive 
and negative 
instances.  {\color{black}We contrast the general ($\bm{o}^{\text{gen}}$) and full ($\bm{o}^{\text{full}}$) representations (as positive and negative instances) for the same input $\bm{x}$ 
to make them different, which 
encourages the learning of domain-specific knowledge in $\bm{o}^{\text{full}}$ 
that is not already in the general knowledge and yet related to and integrated with the general knowledge ($\bm{o}^{\text{gen}}$) of the input. 
}

We construct contrastive instances as follows: for an input $\bm{x}_m$, three contrastive instances are produced. Anchor $\bm{o}_m$ and positive instance $\bm{o}_m^+$ are both full knowledge from  Eq.~\ref{eq:final_specialized_knowledge}, obtained based on two independently sampled dropout masks in the Transformer (recall that this can be achieved by inputting $\bm{x}_m$ twice (see Sec.~\ref{sec: prevent_forgetting}). We regard $\bm{o}_m^+$ and $\bm{o}_m$ as positive instances because the dropout noise has been shown to be good positive instances for improving alignment in training sentence embedding~\cite{gao2021simcse}. Negative instance $\bm{o}_m^-$ is the general knowledge for $\bm{x}_m$ from the LM obtained via Eq.~\ref{eq:final_general_knowledge}. 
}
{\color{black}With $\bm{o}_m$, $\bm{o}_m^+$, and $\bm{o}_m^-$, our contrastive loss is ($\text{sim}(\cdot)$ is the cosine similarity), 
\begin{equation}
\label{eq:contrast}
\small
\mathcal{L}_{\text{contrast}} =  -\text{log}\frac{e^{\text{sim}(\bm{o}_m,\bm{o}_m^+)}/\tau}{\sum_{j=1}^N(e^{\text{sim}(\bm{o}_m,\bm{o}_j^+)/\tau}+e^{\text{sim}(\bm{o}_m,\bm{o}_j^-)/\tau})}.
\end{equation}
Compared to Eq.~\ref{eq.relate_contrast}, the second term is added in the denominator, i.e., general knowledge representations as additional negative samples/instances. Figure~\ref{fig:overview} (B) shows a red arrow pointed from $\bm{o}^{\text{full}}$ to itself, indicating the positive instances are from inputting twice. The dashed red arrow pointing to $\bm{o}^{\text{gen}}$ indicates the negative instances contrasting the specialized and general knowledge.}


\begin{table*}[]
\centering
\resizebox{\textwidth}{!}{
\begin{tabular}{ccc|ccccc}
\specialrule{.2em}{.1em}{.1em}
\multicolumn{3}{c|}{Unlabeled Domain Datasets} & \multicolumn{5}{c}{End-Task Classification Datasets} \\
Source & Dataset & Size & Dataset & Task & \#Training & \#Testing & \#Classes \\
\specialrule{.1em}{.05em}{.05em}
\multirow{3}{*}{Reviews} & Yelp Restaurant & 758MB & Restaurant & Aspect Sentiment Classification (ASC) & 3,452 & 1,120 & 3 \\
 & Amazon Phone & 724MB & Phone & Aspect Sentiment Classification (ASC) &  239 & 553 & 2 \\
 & Amazon Camera & 319MB & Camera & Aspect Sentiment Classification (ASC) & 230 & 626 & 2 \\
 \hline
\multirow{3}{*}{Academic Papers} & ACL Papers & 867MB & ACL & Citation Intent Classification & 1,520 & 421 & 6 \\
 & AI Papers & 507MB & AI & Relation   Classification & 2,260 & 2,388 & 7 \\
 & PubMed Papers & 989MB & PubMed & Chemical-protein Interaction Prediction & 2,667 & 7,398 & 13 \\
\specialrule{.1em}{.05em}{.05em}
\end{tabular}
}
\caption{
Statistics for domain post-training datasets and end task supervised classification datasets (more detail of each task is given in Appendix~\ref{ap:dataset}).  
} 
\label{tab:dataset}
\vspace{-4mm}
\end{table*}

\subsection{DGA Objectives} DGA is a pipelined model: First, {\color{black}a subset of the domain data} is used to estimate the attention head importance ($I_{lh}$ in Sec.~\ref{sec: prevent_forgetting}). Second, given the attention head importance, we compute the final domain-adaptive loss by combining the conventional Masked Language Model (MLM) loss (include the proposed soft-masking for general knowledge) and the proposed contrastive loss:
\begin{equation}
\label{eq:final_loss}
\mathcal{L}_{\text{DA-train}} = \mathcal{L}_{\text{MLM}} + \lambda_1\mathcal{L}_{\text{contrast}},
\end{equation}
where {\color{black}$\lambda_1$ is the hyper-parameter} to adjust the impact of the added term.

\section{Experiments}
\label{sec:experiments}
We follow the experiment setup in \cite{DBLP:conf/acl/GururanganMSLBD20}. 
RoBERTa \cite{DBLP:journals/corr/abs-1907-11692}\footnote{\url{https://huggingface.co/roberta-base}} is used as the LM. In each experiment, we first DA-train the LM and then fine-tune it on the end-task. The final evaluation is based on the end-task results.

\subsection{Datasets and Baselines}
\label{sec.data-baselines}

\textbf{Datasets:} 
Table \ref{tab:dataset} shows the statistics of the \textit{unlabeled domain datasets} for DA-training and their corresponding \textit{end-task classification datasets}. We use 6 \textit{unlabeled domain datasets}:\footnote{We down-sampled the \textit{PubMed} due to its huge original size. In general, our datasets are much smaller comparing to previous work \cite{DBLP:conf/acl/GururanganMSLBD20} (which used more than 11GB of data for each domain). Our experiments showed that a smaller dataset is sufficient and more data does not help. It also requires much less computation resource.} 3 of them are about reviews: \textit{Yelp Restaurant} \cite{DBLP:conf/naacl/XuLSY19}, 
\textit{Amazon Phone} \cite{DBLP:conf/emnlp/NiLM19}, \textit{Amazon Camera} \cite{DBLP:conf/emnlp/NiLM19}; 3 of them are academic papers: \textit{ACL Papers} \cite{DBLP:conf/acl/LoWNKW20}, \textit{AI Papers} \cite{DBLP:conf/acl/LoWNKW20}, and \textit{PubMed Papers}\footnote{\url{https://pubmed.ncbi.nlm.nih.gov/}}.
Each unlabeled domain dataset has a corresponding \textit{end-task classification dataset}\footnote{Note that our results are different from those presented in Table 5 of \cite{DBLP:conf/acl/GururanganMSLBD20} because we observe very high variances due to very small original test sets and thus re-partition the training and test set (by enlarging the test set and reducing the training set slightly)}: \textit{Restaurant}\footnote{To be consistent with existing research \cite{tang-etal-2016-aspect}, examples with conflict polarities (both positive and negative sentiments
are expressed about an aspect term) are not used.} \cite{DBLP:conf/naacl/XuLSY19}, \textit{Phone} \cite{ding2008holistic,hu2004mining}, \textit{Camera} \cite{ding2008holistic,hu2004mining}\footnote{Note that \citet{ding2008holistic} and \citet{hu2004mining} contain 9 and 5 domains, respectively. We extract those domains related to ``Phone'' and ``Camera'' from them.}, \textit{ACL} (ACL-ARC in \citet{DBLP:journals/tacl/JurgensKHMJ18}), \textit{AI} (SCIERC in \citet{DBLP:conf/emnlp/LuanHOH18}), and PubMed (CHEMPORT in \citet{kringelum2016chemprot}).  

\vspace{+1.5mm}
\noindent
\textbf{Baselines.} 
We consider 10 baselines. 

(1).~\textbf{Non-DA-training (RoBERTa)}~\cite{DBLP:journals/corr/abs-1907-11692} uses the original RoBERTa for the end-task fine-tuning without any DA-training. 

(2).~\textbf{DA-training using masked language model loss (MLM)} is the existing DA-training method. To our knowledge, existing DA-training systems are all based on the MLM loss. 

(3).~\textbf{DA-training using adapter-tuning (MLM  (Adapter))} adds adapter layers between layers of Transformer for DA-training. An adapter~\cite{Houlsby2019Parameter} has two fully connected layers and a skip connection. During DA-training, the Transformer is fixed, only the adapters are trained. The bottleneck (adapter) size is set to 64~\cite{Houlsby2019Parameter}. 
During end-task fine-tuning, both RoBERTa and adapters are trainable for fair comparison. 

(4).~\textbf{DA-training using prompt-tuning (MLM (Prompt))}~\cite{DBLP:conf/emnlp/LesterAC21} adds a sequence of prompt tokens to the end of the original sequence. In DA-training, RoBERTa (the LM) is fixed and only the prompt tokens are trained. In end-task fine-tuning, both LM and the trained prompt are trainable. We initialize 100 tokens and set the learning rate of the prompt token to 0.3 in DA-training, following the setting in \citet{DBLP:conf/emnlp/LesterAC21}. 

(5).~\textbf{Knowledge distillation (MLM+KD)}~\cite{hinton2015distilling} minimizes the representational deviation  between the general knowledge in LM and the specialized knowledge in DA-training. We compute the KL divergence between  the representations (the output before the masked language model prediction head) of each word of the two models (LM and DA-trained) as the distillation loss.

(6).~\textbf{Adapted distillation through attention (MLM+AdaptedDeiT)} is derived from DeiT~\cite{touvron2021training}, a distillation method for visual Transformer (ViT) \cite{dosovitskiy2020image}. 
We adapt DeiT to a text-based and unsupervised model by distilling the LM representation\footnote{We take the average of its token's output as sequence's representation. The same for SimCSE baseline.} to the added distillation token and change ViT to RoBERTa. 

{\color{black}(7,~8).~\textbf{DA-training using sequence-level contrastive learning (MLM+SimCSE and MLM+InfoWord)}. SimCSE is a contrastive learning method for sentence embedding~\cite{gao2021simcse}. We use its unsupervised version where positive samples are from the same input with different dropout masks and negative samples are other instances in the same batch. InfoWord~\cite{DBLP:conf/iclr/KongdYLDY20} is another contrastive learning method contrasts the span-level local representation and sequence-level global representation. 

(9,~10).~\textbf{DA-training using token-aware contrastive learning (MLM+TaCL and MLM+TaCO)}. TaCL~\cite{su2021tacl} and TaCO~\cite{DBLP:conf/acl/FuZX0022} are two recent methods to improve BERT pre-training with token-aware contrastive loss.\footnote{TaCL and TaCO are not a DA-training model. It pre-trains an LM to improve it using the same data as that for training the LM. We switch the data to our target domain data.} We change the backbone to RoBERTa for fair comparison.} 



\subsection{Implementation Detail}
\label{sec:imp_detail}
\textbf{Architecture.} We adopt $\text{RoBERTa}_{\textbf{BASE}}$ as our backbone LM (12 layers and 12 attention heads in each layer). A masked language model head is applied for DA-training. The end-task fine-tuning of RoBERTa follows the standard practice. 
For the three ASC tasks (see Table~\ref{tab:dataset}), we adopt the ASC formulation in \cite{DBLP:conf/naacl/XuLSY19}, where the aspect (e.g., ``\textit{sound}'') and review sentence (e.g., ``\textit{The sound is great}'') are concatenated via \texttt{</s>}. 

\textbf{Hyperparameters.} 
Unless otherwise stated, the same hyper-parameters are used in all experiments. The maximum input length is set to 164 which is long enough for all datasets. Adam optimizer is used for both DA-training and end-task fine-tuning. The max sequence length is set to 164, which is long enough for our end-tasks and only needs moderate computational resources. 

\textbf{Domain-adaptive pre-training (DA-training).} The learning rate is set to 1e-4 and batch size is 256. We train 2.5K steps for each domain, roughly a full pass through the domain data, following \cite{DBLP:conf/acl/GururanganMSLBD20,DBLP:conf/naacl/XuLSY19}. The subset of data $\{\bm{x}^{\text{sub}}_m\}$ for computing $\mathcal{L}_{\text{impt}}$ to determine head importance in Sec.~\ref{sec: prevent_forgetting} is set to 1.64 Million tokens, 
which is sufficient in our experiments. $\lambda_1$ in Eq.~\ref{eq:final_loss} is set to 1 and $\tau$ in Eq.~\ref{eq:contrast} is set to 0.05. 

\textbf{End-task fine-tuning.} The learning rate is set to 1e-5 and batch size to 16. We train on end-task fine-tuning datasets for 5 epochs for Restaurant; 10 epochs for ACL, AI and PubMed; and 15 epochs for Phone and Camera. We simply take the results for the last epoch as 
we empirically found that the above number of epochs gives us stable and convergence results.

\begin{table*}[h]
\centering
\resizebox{\textwidth}{!}{
\begin{tabular}{c|cccccccccccc}
\specialrule{.2em}{.1em}{.1em}
Domain & \multicolumn{2}{c}{Camera} & \multicolumn{2}{c}{Phone} & \multicolumn{2}{c}{Resturant} & \multicolumn{2}{c}{AI} & \multicolumn{2}{c}{ACL} & PubMed & \multirow{2}{*}{Avg}\\
Model & MF1 & Acc. & MF1 & Acc. & MF1 & Acc. & MF1 & Acc. & MF1 & Acc. & Micro-F1  \\
\specialrule{.1em}{.05em}{.05em}
RoBERTa & 78.82 & 87.03 & 83.75 & 86.08 & 79.81 & 87.00 & 60.98 & 71.85 & 66.11 & 71.26 & 72.38 & 73.64 \\
\hline
MLM & 84.39 & 89.90 & 82.59 & 85.50 & 80.84 & 87.68 & 68.97 & 75.95 & 68.75 & 73.44 & 72.84 & 76.40 \\
MLM (Adapter) & 83.62 & 89.23 & 82.71 & 85.35 & 80.19 & 87.14 & 60.55 & 71.38 & 68.87 & 72.92 & 71.68 & 74.60 \\
MLM (Prompt) & 85.52 & 90.38 & 84.17 & 86.53 & 79.00 & 86.45 & 61.47 & 72.36 & 66.66 & 71.35 & 73.09 & 74.98 \\
\hline
MLM+KD & 82.79 & 89.30 & 80.08 & 83.33 & 80.40 & 87.25 & 67.76 & 75.46 & 68.19 & 72.73 & 72.35 & 75.26 \\
MLM+AdaptedDeiT & 86.86 & 91.37 & 83.08 & 85.64 & 79.70 & 86.84 & 69.72 & 76.83 & 69.11 & 73.35 & 72.69 & 76.86\\
MLM+SimCSE & 84.91 & 90.35 & 83.46 & 86.08 & 80.88 & 87.59 & 69.10 & 76.25 & 69.89 & 74.30 & 72.77 & 76.84 \\
MLM+TaCL & 81.98 & 88.88 & 81.87 & 84.92 & 81.12 & 87.50 & 64.04 & 73.18 & 63.18 & 70.31 & 69.46  & 73.61\\
MLM+TaCO & 84.50 & 90.22 & 82.63 & 85.32 & 79.27 & 86.68 & 59.73 & 71.22 & 63.66 & 70.36 & 72.38 & 73.69 \\
MLM+InfoWord & 87.95 & 91.92 & 84.58 & 86.84 & 81.24 & 87.82 & 68.29 & 75.92 & 68.58 & 73.68 & 73.21 & 77.31 \\
\hline
DGA & \textbf{88.52} & \textbf{92.49} & \textbf{85.47} & \textbf{87.45} & \textbf{81.83} & \textbf{88.20} & \textbf{71.99} & \textbf{78.06} & \textbf{71.01} & \textbf{74.73} & \textbf{73.65} & \textbf{78.74} \\
\specialrule{.1em}{.05em}{.05em}
\end{tabular}}
\caption{We report the macro-F1 (MF1) and accuracy results for all datasets, except for CHEMPORT in the PubMed domain, for which we use micro-F1 following~\citet{DBLP:conf/acl/GururanganMSLBD20,dery2021should,beltagy-etal-2019-scibert}. The results are averages of 5 random seeds (the standard deviation is reported in Appendix~\ref{ap:std}). {\color{black} The average column (Avg) is the average over the MF1 (or Micro-F1 for PubMed) for all datasets.}
} 
\label{tab:dapt_result}
\end{table*}

\subsection{Evaluation Results and Ablation Study}
\label{sec:results}

{\color{black}
We report the end-task results of the 10 baselines on the 6 datasets in Table~\ref{tab:dapt_result}.

\textbf{Superiority of DGA.} Our DGA consistently outperforms all baselines. Thanks to the proposed more informed  adaptation, DGA improves over the widely used \textit{traditional DA-training baseline} MLM. We also see that MLM markedly outperforms RoBERTa (non-DA-training) on average (see the last column). We discuss more observations about the results bellow.

(1). Training the entire LM in DGA helps achieve much better results. Using adapter (MLM (adapter)) and prompt (MLM (prompt)) have mixed results. This is because adapter and prompt do not have sufficient trainable parameters, which are also randomly initialized and can be difficult to train.

(2). DGA is also better than distillation-based systems: MLM+AdaptedDeiT and MLM+KD, which try to preserve the past knowledge. This is not surprising because the goal of DA-training is not simply preserving the previous knowledge but also to adapt/change it as needed to suit the target domain. DGA is specifically designed for this with soft-masking and contrasting of knowledge. 

(3). The contrastive learning in DGA is more effective than the other contrastive alternatives 
(MLM+SimCSE, MLM+TaCL, MLM+TaCO and MLM+InfoWord). This indicates  contrasting the general and full knowledge for knowledge integration is important. 
} 

\begin{table*}[]
\centering
\resizebox{\textwidth}{!}{
\begin{tabular}{c|cccccccccccc}
\specialrule{.2em}{.1em}{.1em}
Domain & \multicolumn{2}{c}{Camera} & \multicolumn{2}{c}{Phone} & \multicolumn{2}{c}{Resturant} & \multicolumn{2}{c}{AI} & \multicolumn{2}{c}{ACL} & PubMed & \multirow{2}{*}{Avg}\\
Model & MF1 & Acc. & MF1 & Acc. & MF1 & Acc. & MF1 & Acc. & MF1 & Acc. & Micro-F1  \\
\specialrule{.1em}{.05em}{.05em}
RoBERTa & 78.82 & 87.03 & 83.75 & 86.08 & 79.81 & 87.00 & 60.98 & 71.85 & 66.11 & 71.26 & 72.38 & 73.64 \\
\hline
MLM & 84.39 & 89.90 & 82.59 & 85.50 & 80.84 & 87.68 & 68.97 & 75.95 & 68.75 & 73.44 & 72.84  & 76.40\\
\hline
DGA (H, I) & 86.79 & 91.60 & 84.21 & 86.40 & 81.32 & 87.91 & 71.07 & 77.36 & 69.50 & 73.82 & 73.34 & 77.71 \\
DGA (H, I, O) &  88.04 & 92.01 & 85.85 & 87.63 & 81.45 & 87.79 & 71.54 & 77.61 & 70.52 & 74.58 & 73.10 & 78.42 \\
DGA (H, I, O, E) & 87.05 & 91.60 & 83.74 & 86.11 & 80.64 & 87.61 & 72.64 & 78.17 & 71.24 & 74.96 & 73.54 & 78.14 \\
\hline
DGA (w/o contrast) & 86.19 & 90.89 & 84.48 & 86.65 & 81.70 & 87.93 & 68.25 & 75.49 & 69.31 & 73.73 & 72.72 & 77.11\\
DGA (random mask) & 82.07 & 89.30 & 83.86 & 86.33 & 80.60 & 87.52 & 69.51 & 76.64 & 69.59 & 73.73 & 72.92 & 76.43\\
Ensemble (LM+MLM) & 85.22 & 90.64 & 85.15 & 87.23 & 79.86 & 86.98 & 65.10 & 74.43 & 68.56 & 73.44 & 72.60 & 76.08\\
DGA (domain-specific) & 88.06 & 92.04 & 83.45 & 85.82 & 81.72 & 87.90 & 68.00 & 75.57 & 70.91 & 75.06 & 73.17 & 77.55\\
\hline
DGA & \textbf{88.52} & \textbf{92.49} & \textbf{85.47} & \textbf{87.45} & \textbf{81.83} & \textbf{88.20} & \textbf{71.99} & \textbf{78.06} & \textbf{71.01} & \textbf{74.73} & \textbf{73.65} & \textbf{78.74}  \\
\specialrule{.1em}{.05em}{.05em}
\end{tabular}}
\caption{Ablation results - averages of 5 random seeds. The standard deviations are reported in Appendix B.
} 
\vspace{-4mm}
\label{tab:ablation}
\end{table*}

{\color{black}\textbf{Effectiveness of the proxy KL-divergence loss.} We use the proposed \textit{proxy KL-divergence loss} to compute the head importance to identify the general language knowledge in the LM without using the LM's original pre-training data (Sec. \ref{sec: prevent_forgetting}). 

For evaluation, we are interested in how good the proxy is. Since we don’t have the data that pre-trains RoBERTa, it is not obvious how to assess the quality of the proxy directly. Here, we provide some indirect evidences to show the effectiveness of the proxy for computing the importance of units to the general knowledge in the LM. 

 We conduct a separate experiment to compare the attention heads' importance score vectors after applying the proxy using the data from different domains. For each domain $i$, we compare its importance vector with the importance vector of every other domain, and 
then average the cosine similarities to get the value for domain $i$.
We get 0.92 for Restaurant, the same 0.91 for ACL, AI, and Phone, 0.89 for PubMed and 0.92 for Camera. We see that different domains give similar importance values, which indirectly show that our proxy can identify the common general knowledge. 

We also compute the importance score distributions of the proxy. For each of the 6 domains, after applying the proxy, 
around 20\% of the attention heads are heavily protected ($0.8 \le I^{\text{norm}}_{lh} \le 1.0$) and another 20\% moderately protected ($0.6 \le I^{\text{norm}}_{lh} < 0.8$), which indicate the general knowledge. While Phone, AI, Camera and Restaurant share a similar distribution, ACL and PubMed protect slightly less. This is understandable as PubMed and ACL (medical or NLP publications) are probably less common than the other domains and the general knowledge in the LM covers them less.
}

{\color{black}\textbf{Ablation study.} To better understand DGA, We want to know (1) whether constraining the neurons in other layers are helpful (the proposed DGA only constrains the attention heads), and (2) where the gain of DGA is from. 
To answer (1), we constrain the training of different layers in a standard Transformer. In Table \ref{tab:ablation} (rows 3-5), ``\textbf{H}'', ``\textbf{I}'', and ``\textbf{O}'' refer to attention head, intermediate layer, output layer in a standard Transformer layer, respectively. ``\textbf{E}'' refers to the embedding layers. The brackets with combination of ``H, I, O, E'' indicate the location we apply the soft-masking (DGA only applies soft-masking in the attention head). We can see their results are similar or worse than DGA, implying that attention heads are more indicative of important knowledge. To answer (2), we conduct the following ablation experiments: (i) \textbf{DGA (w/o contrast)}, without the contrastive loss, but only soft-masking the backward pass according to the attention head importance. (ii) \textbf{DGA (random masking)} with randomly generated attention head importance scores and using them to do soft-masking and contrastive learning. 
(iii) \textbf{Ensemble (LM+MLM)} performs the end-task fine-tuning on both the MLM DA-trained RoBERTa (conventional DA-training) and the original RoBERTa (LM) by concatenating their outputs and taking the average. (iv) \textbf{DGA (domain-specific)} refers to the variant that contrasts domain-specific and general knowledge (see Sec.~\ref{sec.contrast}).\footnote{We don't have DGA(w/o soft-masking) because our contrastive learning relies on soft-masking. If removed, contrastive loss will not have the additional negative samples and our DGA becomes MLM+SimCSE.}

Table~\ref{tab:ablation} shows that the full DGA always gives the best result, indicating every component contributes. Additional observations are as follows:

(1) 
DGA's gain is partially from the novel soft-masking: we can see that on average, DGA (w/o contrast) outperforms conventional DA-training (MLM). Besides, our gradient-based mask is informative: we can see DGA (random mask) is worse than DGA (w/o contrast) on all datasets. DGA (w/o contrast) is even better than Ensemble, which directly combines the information given by both the original LM and the traditional DA-trained model during end-task fine-tuning

(2) 
Besides soft-masking, contrasting the general and full knowledge also  helps. We can see DGA outperforms DGA (w/o contrast) and DGA (domain-specific) in all datasets.


}

\section{Conclusion}

This paper argued that an effective DA-training method should effectively integrate the target domain knowledge to the general knowledge in the LM. Existing approaches do not explicitly do this. This paper proposed a novel method DGA to achieve it (1) by estimating the attention heads importance in LM and using the importance scores to soft-mask the attention heads in DA-training to preserve the important knowledge in LM as much as possible, and (2) by contrasting the general and the full knowledge. 
Extensive experiment results demonstrated the effectiveness of the proposed approach DGA. 

\section{Limitations}
While effective, DGA has some limitations. First, the main focus of DGA is to adapt an LM to a given target domain. It does not consider the generalization to other domains. For example, it will be interesting to incrementally or continually adapt an LM to more and more domains to make the LM more useful~\cite{ke2020mixed,DBLP:journals/corr/abs-2112-02706,ke2022cpt}. 
Second, the importance of parameters for general knowledge in the LM is computed using a proxy method based on model robustness. Although it is quite effective, it is interesting to explore other approaches to further improve it. We will work on these in our future work as specializing and improving an LM is an important problem. 

\section*{Acknowledgments}
{\color{black}The work of Zixuan Ke and Bing Liu was supported in part by three National Science Foundation (NSF) grants (IIS-1910424, IIS-1838770, and CNS-2225427).} 

\bibliography{anthology,custom}
\bibliographystyle{acl_natbib}

\appendix

\begin{table*}[]
\centering
\resizebox{\textwidth}{!}{
\begin{tabular}{c|ccccccccccc}
\specialrule{.2em}{.1em}{.1em}
Domain & \multicolumn{2}{c}{Camera} & \multicolumn{2}{c}{Phone} & \multicolumn{2}{c}{Resturant} & \multicolumn{2}{c}{AI} & \multicolumn{2}{c}{ACL} & PubMed \\
Model & MF1 & Acc. & MF1 & Acc. & MF1 & Acc. & MF1 & Acc. & MF1 & Acc. & Micro-F1  \\
\specialrule{.1em}{.05em}{.05em}
RoBERTa & $\pm${0.0403} & $\pm${0.0179} & $\pm${0.0210} & $\pm${0.0154} & $\pm${0.0117} & $\pm${0.0049} & $\pm${0.0646} & $\pm${0.0347} & $\pm${0.0192} & $\pm${0.0096} & $\pm${0.0071}  \\
\specialrule{.1em}{.05em}{.05em}
MLM & $\pm${0.0479} & $\pm${0.0298} & $\pm${0.0165} & $\pm${0.0103} & $\pm${0.0096} & $\pm${0.0056} & $\pm${0.0117} & $\pm${0.0086} & $\pm${0.0218} & $\pm${0.0118} & $\pm${0.0035} \\
MLM (adapter) & $\pm${0.0165} & $\pm${0.0110} & $\pm${0.0265} & $\pm${0.0181} & $\pm${0.0102} & $\pm${0.0068} & $\pm${0.0551} & $\pm${0.0288} & $\pm${0.0142} & $\pm${0.0099} & $\pm${0.0055} \\
MLM (prompt) & $\pm${0.0243} & $\pm${0.0138} & $\pm${0.0126} & $\pm${0.0087} & $\pm${0.0060} & $\pm${0.0035} & $\pm${0.0301} & $\pm${0.0124} & $\pm${0.0068} & $\pm${0.0108} & $\pm${0.0028} \\
MLM+KD & $\pm${0.0295} & $\pm${0.0158} & $\pm${0.0320} & $\pm${0.0230} & $\pm${0.0099} & $\pm${0.0070} & $\pm${0.0345} & $\pm${0.0224} & $\pm${0.0292} & $\pm${0.0155} & $\pm${0.0093} \\
MLM+AdaptedDeiT & $\pm${0.0187} & $\pm${0.0122} & $\pm${0.0160} & $\pm${0.0101} & $\pm${0.0048} & $\pm${0.0022} & $\pm${0.0250} & $\pm${0.0179} & $\pm${0.0065} & $\pm${0.0079} & $\pm${0.0086} \\
MLM+SimCSE & $\pm${0.0114} & $\pm${0.0077} & $\pm${0.0098} & $\pm${0.0065} & $\pm${0.0029} & $\pm${0.0016} & $\pm${0.0086} & $\pm${0.0056} & $\pm${0.0054} & $\pm${0.0071} & $\pm${0.0027} \\
MLM+TaCL & $\pm${0.0218} & $\pm${0.0103} & $\pm${0.0230} & $\pm${0.0159} & $\pm${0.0105} & $\pm${0.0059} & $\pm${0.0275} & $\pm${0.0156} & $\pm${0.0713} & $\pm${0.0394} & $\pm${0.0118} \\
MLM+TaCO & $\pm${0.0456} & $\pm${0.0232} & $\pm${0.0166} & $\pm${0.0134} & $\pm${0.0077} & $\pm${0.0052} & $\pm${0.0675} & $\pm${0.0380} & $\pm${0.0207} & $\pm${0.0128} & $\pm${0.0099} \\
MLM+InfoWord & $\pm${0.0267} & $\pm${0.0139} & $\pm${0.0272} & $\pm${0.0191} & $\pm${0.0170} & $\pm${0.0089} & $\pm${0.0344} & $\pm${0.0219} & $\pm${0.0070} & $\pm${0.0079} & $\pm${0.0072} \\
DGA & $\pm${0.0095} & $\pm${0.0047} & $\pm${0.0127} & $\pm${0.0094} & $\pm${0.0052} & $\pm${0.0040} & $\pm${0.0127} & $\pm${0.0081} & $\pm${0.0079} & $\pm${0.0080} & $\pm${0.0034} \\
\specialrule{.1em}{.05em}{.05em}
\end{tabular}}
\caption{Standard deviations of the corresponding metrics of the proposed
DGA model and the baselines on the six experiments.
} 
\label{tab:dapt_std}
\end{table*}

\begin{table*}[]
\centering
\resizebox{\textwidth}{!}{
\begin{tabular}{c|ccccccccccc}
\specialrule{.2em}{.1em}{.1em}
Domain & \multicolumn{2}{c}{Camera} & \multicolumn{2}{c}{Phone} & \multicolumn{2}{c}{Resturant} & \multicolumn{2}{c}{AI} & \multicolumn{2}{c}{ACL} & PubMed \\
Model & MF1 & Acc. & MF1 & Acc. & MF1 & Acc. & MF1 & Acc. & MF1 & Acc. & Micro-F1  \\
\specialrule{.1em}{.05em}{.05em}
RoBERTa & $\pm${0.0403} & $\pm${0.0179} & $\pm${0.0210} & $\pm${0.0154} & $\pm${0.0117} & $\pm${0.0049} & $\pm${0.0646} & $\pm${0.0347} & $\pm${0.0192} & $\pm${0.0096} & $\pm${0.0071}  \\
\specialrule{.1em}{.05em}{.05em}
MLM & $\pm${0.0479} & $\pm${0.0298} & $\pm${0.0165} & $\pm${0.0103} & $\pm${0.0096} & $\pm${0.0056} & $\pm${0.0117} & $\pm${0.0086} & $\pm${0.0218} & $\pm${0.0118} & $\pm${0.0035} \\
\hline
DGA (H, I) & $\pm${0.0373} & $\pm${0.0210} & $\pm${0.0032} & $\pm${0.0039} & $\pm${0.0054} & $\pm${0.0045} & $\pm${0.0095} & $\pm${0.0048} & $\pm${0.0094} & $\pm${0.0073} & $\pm${0.0049}  \\
DGA (H, I, O) & $\pm${0.0167} & $\pm${0.0092} & $\pm${0.0182} & $\pm${0.0155} & $\pm${0.0055} & $\pm${0.0033} & $\pm${0.0093} & $\pm${0.0075} & $\pm${0.0080} & $\pm${0.0070} & $\pm${0.0056} \\
DGA (H, I, O, E) & $\pm${0.0237} & $\pm${0.0123} & $\pm${0.0270} & $\pm${0.0187} & $\pm${0.0099} & $\pm${0.0050} & $\pm${0.0109} & $\pm${0.0089} & $\pm${0.0067} & $\pm${0.0057} & $\pm${0.0079}  \\
\hline
DGA (w/o contrast) & $\pm${0.0433} & $\pm${0.0251} & $\pm${0.0135} & $\pm${0.0106} & $\pm${0.0060} & $\pm${0.0040} & $\pm${0.0197} & $\pm${0.0119} & $\pm${0.0132} & $\pm${0.0093} & $\pm${0.0050} \\
DGA (random mask) & $\pm${0.0879} & $\pm${0.0413} & $\pm${0.0335} & $\pm${0.0235} & $\pm${0.0096} & $\pm${0.0044} & $\pm${0.0153} & $\pm${0.0090} & $\pm${0.0105} & $\pm${0.0059} & $\pm${0.0052}  \\
Ensemble & $\pm${0.0332} & $\pm${0.0178} & $\pm${0.0199} & $\pm${0.0139} & $\pm${0.0035} & $\pm${0.0031} & $\pm${0.0236} & $\pm${0.0103} & $\pm${0.0061} & $\pm${0.0028} & $\pm${0.0046} \\
\hline
DGA (domain-specific) & $\pm${0.0137} & $\pm${0.0070} & $\pm${0.0259} & $\pm${0.0200} & $\pm${0.0031} & $\pm${0.0018} & $\pm${0.0128} & $\pm${0.0071} & $\pm${0.0108} & $\pm${0.0067} & $\pm${0.0043}\\
\hline
DGA & $\pm${0.0095} & $\pm${0.0047} & $\pm${0.0127} & $\pm${0.0094} & $\pm${0.0052} & $\pm${0.0040} & $\pm${0.0127} & $\pm${0.0081} & $\pm${0.0079} & $\pm${0.0080} & $\pm${0.0034} \\
\specialrule{.1em}{.05em}{.05em}
\end{tabular}}
\caption{Standard deviations of the corresponding metrics of the proposed
DGA model and the ablation on the six experiments.
} 
\label{tab:ablation_std}
\vspace{-3mm}
\end{table*}

\newpage

\section{Datasets Details}
\label{ap:dataset}
Table 2 in the main paper has given the number of examples in each dataset. Here we provide additional details about the 4 types of end-tasks.

(1) \textbf{(Phone, Camera and Restaurant) Aspect Sentiment Classification (ASC)} is defined as follows~\cite{liu2015sentiment}: given an aspect or product feature (e.g., \textit{picture quality} in a camera review) and a review sentence containing the aspect in a domain or product category (e.g., camera), classify if the sentence expresses a positive, negative, or neutral (no opinion) sentiment or polarity about the aspect (for Phone and Camera, there are only negative and positive polarities in the data).

(2) \textbf{(ACL) Citation Intent Classification} is defined as follows: given a citing sentence (a sentence contains a citation), classify if the sentence expresses a citation function among ``background'', ``motivation'', ``uses'', ``extension'' and ``comparison or contrast future''.

(3) \textbf{(AI) Relation Classification} is defined as follows: given a within-sentence word sequence spans containing a pair of entities, classify if the span expresses a relation among ``feature of'', ``conjunction'', ``evaluate for'', ``hyponym of'', ``used for'', ``part of'' and ``compare''.

(4) \textbf{(PubMed) Chemical-protein Interaction Classification} is defined as follows: given a span containing a pair of chemical and protein, classify if the span expresses a chemical-protein interaction among ``downregulator'', ``substrate'', ``indirect-upregulator'', ``indirect-downregulator'', ``agnonist'', ``activator'', ``product of'', ``agonist-activator'', ``inhibitor'', ``upregulator'', ``substrate product of'', ``agonist-inhibitor''and ``antagonist''.

\section{Standard Deviations}
\label{ap:std}


Table~\ref{tab:dapt_std} reports the standard deviations of the corresponding results in Table~\ref{tab:dapt_result} (in the main paper) of DGA and the considered baselines over 5 runs with random seeds. We can see the results of DGA are stable. Some baselines (e.g., RoBERTa in AI, MLM in Camera and MLM+TaCL in ACL) can have quite large standard deviations. 

Table~\ref{tab:ablation_std} reports the standard deviations of the corresponding results in Table~\ref{tab:ablation} (in the main paper) of DGA and the considered baselines over 5 runs with random seeds. We can see the results of DGA are stable. Some baselines (e.g., DGA (random mask) and DGA (w/o contrast) in Camera) can have quite large standard deviations. 

\end{document}